\documentclass[journal,transmag]{IEEEtran}
\usepackage{amsmath}
\usepackage{cite}
\usepackage{booktabs}
\usepackage{graphicx}
\usepackage{hyperref} 
\usepackage{footnote}
\usepackage{multirow} 
\ifCLASSINFOpdf
\else
\fi

\begin{document}
%
\title{CNNs-based Acoustic Scene Classification using \\ Multi-Spectrogram Fusion and Label Expansions}


\author{
\IEEEauthorblockN{Weiping Zheng\IEEEauthorrefmark{1},
Zhenyao Mo\IEEEauthorrefmark{1},
Xiaotao Xing\IEEEauthorrefmark{1}, and
Gansen Zhao\IEEEauthorrefmark{1}}
\IEEEauthorblockA{\IEEEauthorrefmark{1}School of Computer, South China Normal University,\\ Guangzhou 510631, China \\(e-mail: zhengweiping@m.scnu.edu.cn;mozhenyaomyz@163.com;\\ 20142100007@m.scnu. edu.com; gzhao@m.scnu.edu.cn)}
\thanks{This work was partially supported by the Guangdong Provincial Scientific and Technological Projects of China under Grant 2016B010109005 and the Characteristic Innovation Projects of the Educational Commission of Guangdong Province, China under Grant 2016KTSCX025.}}

\markboth{Journal of \LaTeX\ Class Files,~Vol.~14, No.~8, August~2015}%
{Shell \MakeLowercase{\textit{et al.}}: Bare Demo of IEEEtran.cls for IEEE Transactions on Magnetics Journals}
%



\IEEEtitleabstractindextext{%
\begin{abstract}
Spectrograms have been widely used in Convolutional Neural Networks based schemes for acoustic scene classification, such as the STFT spectrogram and the MFCC spectrogram, etc.  They have different time-frequency characteristics, contributing to their own advantages and disadvantages in recognizing acoustic scenes. In this letter, a novel multi-spectrogram fusion framework is proposed, making the spectrograms complement each other. In the framework, a single CNN architecture is applied onto multiple spectrograms for feature extraction. The deep features extracted from multiple spectrograms are then fused to discriminate the acoustic scenes. Moreover, motivated by the inter-class similarities in acoustic scene datasets, a label expansion method is further proposed in which super-class labels are constructed upon the original classes. On the help of the expanded labels, the CNN models are transformed into the multitask learning form to improve the acoustic scene classification by appending the auxiliary task of super-class classification. To verify the effectiveness of the proposed methods, intensive experiments have been performed on the DCASE2017 and the LITIS Rouen datasets. Experimental results show that the proposed method can achieve promising accuracies on both datasets. Specifically, accuracies of 0.9744, 0.8865 and 0.7778 are obtained for the LITIS Rouen dataset, the DCASE Development set and Evaluation set respectively.
\end{abstract}

}

\maketitle

\IEEEdisplaynontitleabstractindextext

%
\IEEEpeerreviewmaketitle

\section{Introduction}
%
%
%
%
\IEEEPARstart{A}{coustic} scene classification (ASC)~\cite{Barchiesi2015Acoustic} has shown huge potentials in many industrial and business applications. It has gained increasing attentions in these years. Recently, many convolutional neural networks (CNNs) based solutions have been proposed for acoustic scene classification~\cite{Yu2017Recent}. They have shown promising performances. Hamid et al.~\cite{Eghbal2016CP} proposed a hybrid approach using deep CNN and binaural i-vectors~\cite{Dehak2011Front}. This work ranked first in the DCASE2016 challenge. Soo et al.~\cite{Bae2016Acoustic} presented a structure composed of parallel LSTM~\cite{Graves1997Long} and CNN networks which aimed to extract both sequential and spectro-temporal locality information. Using the similar combined structure with~\cite{Bae2016Acoustic} as the feature extractor, Seongkyu et al.~\cite{mun2017generative} proposed a GAN-based~\cite{Goodfellow2014Generative} data augmentation method for ASC and won the first prize in the DCASE2017 challenge. \\
\indent In the existing works, CNNs are often combined with other deep networks (such as LSTM~\cite{Bae2016Acoustic}, GRNN~\cite{Ren2018Deep,Chung2014Empirical}) or traditional features (such as i-vector~\cite{Eghbal2016CP}). In this letter, we propose a multi-spectrogram fusion framework using a single CNN architecture. The CNN network is applied onto multiple spectrograms for feature extraction, resulting in a simple system architecture. Experiments have demonstrated satisfactory results of the proposed methods. On the other hand, inter-class similarity is getting more prominent in ASC. Scene labels are commonly notated according to the functions of the surrounding spaces where the audio segments are recorded. As a result, there are audio segments which are very similar in acoustic properties while assigned as different scenes, for example, the audio segments of the libraries and that of the bookstores. However, they are simply considered as different classes in the training process. Ignoring their similarities in acoustic properties will probably prevent the deep networks from learning more essential acoustic features. To distinguish the degrees of similarities among classes, a common solution is to use triple loss~\cite{Zhang2016Embedding} or quadruplet loss~\cite{Han2017Breast} where the hierarchical relations among classes should be carefully organized. Nevertheless, the selections of good anchor points in this method are challenging. Moreover, it requires more training samples and results in a more complex optimization. Multitask learning is another solution, which requires hierarchical labels. However, most acoustic scene datasets merely provide one-level labels. \\
\indent Motivated by above observations, we build up a multi-spectrogram fusion framework where a single CNN is used for feature extraction. Under this framework, we further propose a label expansion method by constructing super-class labels upon the original classes. Using the expanded labels, the building block CNN is modified into multitask learning form accordingly to exploit the inter-class similarities mentioned above. To the best of our knowledge, constructing labels purposely and integrating them into multitask learning to boosting the classification performance have not yet been explored in ASC.\\
\indent In this letter, we produce multiple spectrograms for a single audio segment using different signal processing methods. Specifically, we use the STFT~\cite{Nawab1988Short}, CQT~\cite{Brown1992An}, and MFCC~\cite{Logan2000Mel} spectrograms. The combination of STFT and CQT spectrograms in the fusion scheme has achieved the best performances. We further propose a novel CNN architecture which is trained on these spectrograms respectively. As a result, multiple basic CNN-based models are obtained. These models serve as feature extractors in this fusion scheme. To train the CNN models, spectrograms are split into multiple samples. The samples originated from the same audio segment are fed into a basic model for feature extraction. The extracted deep features are then concatenated using a random sequence. Furthermore, the dimensionality of concatenated feature is reduced by PCA. The resultant feature is called a "global feature" of the audio segment. Corresponding to multiple basic models, multiple global features can be generated for an audio segment. They are further concatenated into an "aggregated feature". Using aggregated features of the audio segments in the training set, a SVM classifier is trained to discriminate the final acoustic scene class. Note that the CNN model is a building block in the proposed framework. It can be replaced by any popular CNN architecture, such as ResNet~\cite{He2015Deep}, and GoogleNet~\cite{Szegedy2014Going} etc. Moreover, on the basis of the outputs of the basic classifiers, we construct super-class labels for the samples using spectral clustering~\cite{Ng2001On}. These expanded labels are used to improve the basic models by transforming them into multitask learning paradigms. By using these updated models in the feature extraction process, the SVM classifier can be updated accordingly. The contributions of this letter are threefold:
\begin{IEEEenumerate}
	\item We propose a multi-spectrogram fusion framework using a single CNN architecture in feature extractions. By aggregating the segment-level features from multiple spectrograms, the proposed framework significantly improves the performance for ASC.
	\item We propose a novel label expansion method which constructs super-class labels by taking advantage of the similarities in acoustic properties among original classes. The constructed hierarchical labels reflect the relations of the original acoustic scenes from the view of the basic classifiers.
	\item We use the artificial hierarchical labels to transform the basic CNN-based models into multitask learning architectures. Moreover, the relations between super-class and included original classes are modeled as regularization constrains in the loss function. In this way, the models are guided to extract more essential acoustic features.
\end{IEEEenumerate}
\section{PROPOSED METHOD}
\label{sec:2}
\begin{figure}[h]
	\centering
	\includegraphics[scale = 0.5]{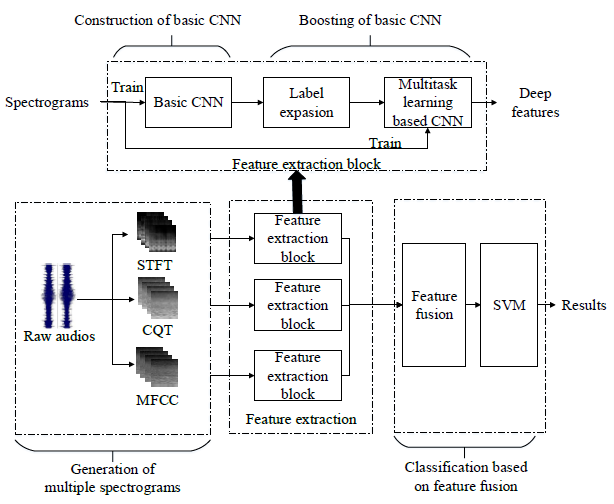}
	\caption{Framework of the proposed multi-spectrogram fusion model.}
	\label{fig：framework}
\end{figure}

	CNN models are usually used in ASC~\cite{Hershey2017CNN} when spectrograms are generated from the audio signal. The spectrogram can be considered as the time-frequency representation of acoustic scene~\cite{Sejdi2009Time,Boashash2015Time}. Since CNN is good at learning spatially local correlations from the input image, it can make good use of the spatial and temporal information in the spectrograms. However, there is a trade-off between frequency and time resolution when producing spectrograms. For some acoustic scene classes with greater temporal recurrent structures, high temporal resolution is more suitable, which requires a smaller frame length. However, some classes prefer high frequency resolution. The preference of resolution will affect the choice of filter shape in CNN model~\cite{Battaglino2016Acoustic}. On the other hand, the resolution is different on the frequency ranges for different time-frequency representations. For example, Constant-Q-Transform (CQT) captures low and mid-to-low frequencies better than the Mel scale [20]. According to above observations, we propose a single CNN model based fusion framework which extract deep features from multiple spectrograms. As seen in Fig.~\ref{fig：framework}, the framework is composed of four parts, including the generation of multiple spectrograms, the construction of basic CNN models, the boosting of CNN models and the classification based on feature fusion. 
	
\section{EXPERIMENTS AND RESULTS}
\label{sec:3}
\subsection{Generating STFT, CQT and MFCC spectrograms}
STFT, CQT and MFCC spectrograms are applied in this letter. Note that other spectrograms can be adopted as well in our framework. When generating spectrograms, audio signals are divided into frames and frame-level frequency representations are produced using certain transforming techniques. By stacking up the frequency representations frame-by-frame, a corresponding spectrogram is obtained. For STFT spectrogram, we perform Short Time Fourier Transform directly on each audio frame. However, Constant-Q-Transform is used in CQT spectrogram which requires more sampling points on the lower frequency range. For MFCC spectrogram, Mel-filter banks are applied and DCT is performed to get cepstral features. After the spectrograms are ready, they are further split into training/testing samples.

\subsection{Constructing basic CNN models for spectrograms}
In our framework, we build up a CNN architecture and train it into multiple classification models. Specifically, we propose a VGG-like network and train it on STFT, CQT and MFCC samples respectively. As a result, three basic CNN models are constructed, namely the VGG-STFT, the VGG-CQT, and the
\begin{table*}[!htbp]
	\centering
	\caption{Architecture of the proposed CNN network}
	\begin{tabular}{lp{0.6cm}p{0.6cm}p{1.2cm}p{0.6cm}p{0.6cm}p{1.2cm}p{0.6cm}p{0.6cm}p{0.6cm}p{0.6cm}p{1.2cm}p{0.6cm}p{0.6cm}p{0.6cm}p{0.6cm}}
		\toprule
		\textbf{Layer} & \textbf{Conv1} & \textbf{Conv2} & \textbf{Pool1} & \textbf{Conv3} & \textbf{Conv4} & \textbf{Pool2} & \textbf{Conv5} & \textbf{Conv6} & \textbf{Conv7} & \textbf{Conv8} & \textbf{Pool3} & \textbf{Conv9} & \textbf{Conv10} & \textbf{Conv11} & \textbf{Full1} \\
		\midrule
		\textbf{Kernel} & 5$\times$5   & 3$\times$3   & Max, 2$\times$2 & 3$\times$3   & 3$\times$3   & Max, 2$\times$2 & 3$\times$3   & 3$\times$3   & 3$\times$3   & 3$\times$3   & Max, 2$\times$2 & 3$\times$3 & 1$\times$1 & 1$\times$1   & - \\
		\midrule
		\textbf{Stride} & 2 & 1 & 2     & 1 & 1 & 2     & 1 & 1 & 1 & 1 & 2     & 1     & 1     & 1 & - \\
		\midrule
		\textbf{Pad} & 2 & 1 & 0   & 1 & 1 & 0     & 1 & 1 & 1 & 1 & 0     & 0     & 0     & 0 & - \\
		\midrule
		\textbf{Channel} & 32 & 32 & 32    & 64 & 64 & 64    & 128 & 128 & 128 & 128 & 128   & 512   & 512   & \textit{M*} & \textit{M*} \\
		\midrule
		\textbf{Activator} & ReLu  & ReLu  & - & ReLu  & ReLu  & - & ReLu  & ReLu  & ReLu  & ReLu  & - & ReLu & ReLu & ReLu  & − \\
		\midrule
		\textbf{BN} & Yes   & Yes   & - & Yes   & Yes   & - & Yes   & Yes   & Yes   & Yes   & - & Yes & Yes & Yes   & − \\
		\midrule
		\textbf{Dropout} & -     & -     & 0.3   & -     & -     & 0.3   & -     & -     & -     & -     & 0.3   & 0.5   & 0.5   & -     & - \\
		\bottomrule
		\multicolumn{10}{l}{$^*$ The variable M represents the number of the output scene classes.}
	\end{tabular}%
	\label{tab:1}%
\end{table*}%
 
\noindent VGG-MFCC models. The architecture of the proposed CNN network is shown in Table~\ref{tab:1}. It was used in our submission to the DCASE2017 challenge~\cite{weiping2017acoustic} and inspired by the one in~\cite{Eghbal2016CP}. To evaluate this network, we compare it with the ResNet in the experiments.\\
\indent For the better description of the methods, it is necessary to introduce some notations and terminologies. Without loss of generality, the architecture of CNN and the spectrogram are not specified below. Suppose we are given a set of n training samples $ S_t=\{(x_1,y_1^o ),\dots,(x_n,y_n^o)\}$ with $y_i^o\in\{1,…,C\} $ indicating the acoustic scene class label of image $ x_i $ (namely a spectrogram patch); superscript o denotes original label of the dataset. For the CNN network, the output layer has C nodes and employs the softmax activation function. Let L be the number of nodes in the next-to-last layer, which is fully connected to the output layer. Let$  W_{j,i} (j\in[1,C];i\in[1,L]) $ denote the weights of connections between these two layers. A negative log-likelihood loss is adopted in the basic CNN model, i.e.,
\begin{equation}
L(W)=\frac{1}{n}\sum_{(x_i,y_i^o)\in S_t}(-\log P(y_i^o|x_i,W))+\alpha||W||_2^2       
\end{equation}

\subsection{Boosting basic CNN models through label expansion}
We propose a label expansion method in this letter. For a certain basic CNN model, such as the VGG-STFT, a confusion matrix $ F $ can be calculated, with $ F_{j,i} $ denotes the number of the samples of class $ j $ which are classified as class $ i $ by that model. To ensure symmetry of the matrix, $ F $ is transformed to a distance matrix $ D $:
\begin{equation}
D=(F+F^T)/2     
\end{equation}
By applying spectral clustering~\cite{Ng2001On} on the matrix $ D $, we divide the original $ C $ classes into $ N $ subsets $ \{H_1,\dots,H_N\} , H_1\cup \dots\cup H_N=\{1,\dots,C\}; H_i\cap H_j=\emptyset ( i\neq j;i,j\in[1,N])$. Each subset is assigned a super-class label. The label space is expanded by adding $ N $ super-class labels into it. Now, every sample has two labels. The $ S_t $ can be rewritten as $ S_t=\{(x_1,<y_1^o,y_1^e>),\dots,(x_n,<y_n^o,y_n^e>)\} $ with $ y_i^e\in\{1,…,N\} $ indicating the super-class label of $ x_i $; superscript $ e $ denotes the expanded label. We have:
\begin{equation}
\forall i\forall j\exists m((y_i^o  \in H_m\wedge y_j^o \in H_m)\to y_i^e=y_j^e)     
\end{equation}
In this method, the degree of similarity between classes is indicated by the number of misclassification. Samples from the same super-class are considered as more similar from the view of the basic model. To make the model aware of the inter-class similarities, we add another output layer onto the basic CNN model, leaving all other details of the model unchanged. The newly added layer has $ N $ output nodes and is fully connected onto the original next-to-last layer. The weights of the newly added connections are denoted as $ U_(j,i) (j\in[1,N];i\in[1,L]) $. We then update the reconstruction error of the new model into the type of multitask learning as follows:
\begin{equation}
\label{equ:4}
\begin{split}
E=\sum_{((x_i,<y_i^o,y_i^e>)\in S_t)}(-\gamma \log P(y_i^o│x_i,W;U)-\\
(1-\gamma)logP(y_i^e│x_i,W;U))
\end{split}
\end{equation}
Considering that the weight vector $ W_j=(W_{j,1},\dots,W_{j,L}) $ for original class $ j $ should capture some similar high-level patterns~\cite{Xie2015Hyper,Gopal2013Recursive} as the weight vector $ U_{s(j)}=(U_{s(j),1},\dots,U_{s(j),L}) $ for the super-class $ s(j) $ of class $ j $, we introduce the following regularization into the loss function:
\begin{equation}
R=\sum_{j=1}^{C}||W_j-U_{s(j)}||_2^2   
\end{equation}
Finally, the loss function of the new model can be defined as:
\begin{equation}
L_{LE} (W;U)=E/n+\alpha R+\beta||W;U||_2^2         
\end{equation}
where $ \alpha $ and $ \beta $ are set to 0.0001, and $ \gamma $ in Eq.~\ref{equ:4} is set to 0.6. After performing this boosting method respectively, we get three new models, namely the VGG-STFT-LE, the VGG-CQT-LE, and the VGG-MFCC-LE models (LE means label expansion is applied).

\subsection{Classifying the acoustic scenes by feature fusion}
The new models are used as feature extractors rather than classifiers.  For a STFT-spectrogram sample v, given a model, say VGG-STFT-LE, the outputs of activation functions of the next-to-last layer in the model are extracted as its deep features, which is denoted as $ f(v) $. As mentioned above, an audio segment g can produce m STFT samples $ [v_1^g,\dots,v_m^g] $, which are split from the STFT spectrogram. The subscripts $ (1,\dots,m) $ present the occurrence order of the samples within the spectrogram. We feed them into the model and extract their corresponding deep features $ f(v_i^g)(i\in[1,m]) $. As the time span of a single sample is short, single deep feature may not depict the whole auditory scene well. To achieve more comprehensive features, they are concatenated into a long feature in random order:
\begin{equation}
O^{STFT} {(g)}=[f(v_{t1}^g );f(v_{t2}^g );\dots;f(v_{tm}^g)]  
\end{equation}
where $ (t1,t2,\dots,tm) $ is a random sequence of numbers $ 1 $ to $ m $. To control the risk of overfitting, the principal component analysis (PCA) method is applied to reduce the dimensionality. Eventually, the global feature of STFT-style for audio segment $ g $ can be achieved. It is denoted as follows:
\begin{equation}
G^{STFT} {(g)}=\phi(O^{STFT} {(g)})    
\end{equation}
where $ \phi(∙) $ means PCA transformation. Likewise, using the CQT and MFCC samples respectively, the global features of $ G^{CQT}(g) $ and  $ G^{MFCC}(g) $ can be obtained as well. \\
\indent These global features are further fused to generate aggregated features. For example, three kinds of aggregated features can be generated here, which are denoted as $ A^{STFT+MFCC}(g) $, $ A^{STFT+CQT}(g) $, and $ A^{MFCC+CQT}(g) $ respectively. Specifically,
\begin{eqnarray}
A^{(STFT+MFCC)}(g)=[G^{STFT}(g);G^{MFCC}(g)] \\
A^{(MFCC+CQT)}(g)=[G^{MFCC}(g);G^{CQT}(g)]   \\
A^{(STFT+CQT)}(g)=[G^{STFT}(g);G^{CQT}(g)]   
\end{eqnarray} 
\indent With these aggregated features, we train SVM classifiers to discriminate the acoustic scenes. Depending on the features used, we can get three SVM classifiers, namely $ SVM^{STFT+CQT}$, $SVM^{MFCC+CQT}$, and $ SVM^{MFCC+CQT}$ respectively. In our experiments, the $SVM^{STFT+CQT}$ achieves the best performances.

\section{EXPERIMENTS AND RESULTS}
\subsection{Experiment Setup}
Two widely used benchmark datasets for ASC are selected to evaluate the performances of our methods. The first one is the DCASE2017 dataset\footnote{http://www.cs.tut.fi/sgn/arg/dcase2017/challenge/download}, which includes two parts, the Development set and the Evaluation set. The second one is the LITIS Rouen dataset~\cite{Rakotomamonjy2017Histogram}\footnote{https://sites.google.com/site/alainrakotomamonjy/home/audio-scene}.\\
\indent We generate 16 STFT-spectrogram patches for each audio segment in DCASE2017 dataset and 12 STFT-spectrogram patches for each in LITIS Rouen dataset. The window size is set to 706 and the hop length is 430.  When dividing the spectrogram into patches, the width is set to 143 and the shift step is 126. The CQT spectrograms are generated by Librosa 0.5.0 using default settings. For DCASE2017 dataset, 20 CQT patches are produced for each audio segment while 16 patches for each in LITIS Rouen dataset. The width is 143 and shift step is 80 in the CQT patches division process. We use a 180-dim feature to generate MFCC spectrogram. It includes 60 static Mel-frequency cepstral coefficients and their first and second derivatives. When performing transformation, the audio segment is divided into 200-ms frames with a hop length of 100 ms. Again, the MFCC spectrogram is divided into patches by the width of 143 and the shift step of 100. Consequently, for each audio segment, we generate 20 MFCC patches in DCASE2017 dataset and 30 patches in LITIS Rouen dataset. For all the patches, we resize them into 143*143 before feeding them into the CNN model.\\
\indent All experiments in this letter are implemented using TensorFlow~\cite{Abadi2015TensorFlow}. During the training of the models, a mini-batch size of 256 is used as well as an early stop strategy with the patience parameter set to 30 and the maximum epoch set to 500. For the VGG-like model, we use Adam~\cite{Kingma2014Adam} as the optimizer with a learning rate of 0.0001. For the ResNet, a momentum optimizer is applied instead, with a learning rate of 0.1 and a momentum of 0.9. In the fusion step, we use the sklearn package in python to implement the PCA and SVM operations. The parameter of n\_components is set to 0.99 for keeping 99\% information after dimension reduction. For the SVM, the linear kernel is applied and the penalty coefficient is set to 1. 

\subsection{Evaluations on DCASE2017 dataset}
On the DCASE2017 dataset, we have implemented two set of experiments. One is totally performed on the Development set, evaluated by the 4-fold cross validation (see column Devel-Set in Table~\ref*{tab:2}). The other one is trained on the Development set and tested on the Evaluation set (see column Eval-Set in Table~\ref{tab:2}). We have generated STFT, CQT and MFCC spectrograms and trained basic classification models respectively. To evaluate our methods, we also use RestNet-50 to build up basic models besides the proposed VGG-like architecture. For each model, we provide two accuracies: one is the sample-level average accuracy and the other is the segment-level majority voting accuracy.\\
\indent As shown in Table~\ref{tab:2}, for the Development set, the best sample-level accuracy is 0.7582; the best voting accuracy is 0.8336, both achieved by the VGG-STFT-LE model. For the Evaluation set, the best sample-level accuracy is 0.5921 by ResNet-CQT-LE, however, the best voting accuracy is 0.684, achieved by ResNet-CQT. From Table~\ref{tab:2}, we can see that all voting accuracies significantly outperform the corresponding sample-level ones. The VGG-like is better in the Development set while the ResNet wins in the Evaluation set. For almost all cases, the performances are improved when label expansion is applied, except the voting accuracy of ResNet-CQT-LE on the Evaluation set. Table~\ref{tab:3} has shown the superiorities of multi-spectrogram fusions. The best accuracy for the Development set is improved from 0.8336 to 0.8865. Meanwhile, the one for the Evaluation set is improved from 0.6840 to 0.7778. The fusions of STFT and CQT achieve the best results for all the models and both datasets. Their fusions can obviously improve the accuracies. Although the best result for Development set is achieved by VGG-LE with STFT+CQT features, ResNet-LE has won in all other cases. Therefore, ResNet is more suitable for DCASE2017. Finally, label expansion is indeed helpful here. It improves the performances for both models with all aggregated features.

\begin{table}[!htbp]
	\centering
	\caption{Classification performances on the DCASE2017 datasets using different single spectrogram.}
	\begin{tabular}{ccccc}
		\toprule
		\multirow{2}[3]{*}{\textbf{Model}} & \multicolumn{2}{c}{\textbf{Devel-Set}} & \multicolumn{2}{c}{\textbf{Eval-Set}} \\
		\cmidrule{2-5}    \multicolumn{1}{c}{} & \multicolumn{1}{c}{\textbf{Sample-Level}} & \multicolumn{1}{c}{\textbf{Voting}} & \multicolumn{1}{c}{\textbf{Sample-Level}} & \multicolumn{1}{c}{\textbf{Voting}} \\
		\midrule
		VGG-STFT & 0.7274 & 0.8056 & 0.4562 & 0.5296 \\
		\midrule
		VGG-STFT-LE & \textbf{0.7582} & \textbf{0.8336} & 0.5356 & 0.6494 \\
		\midrule
		VGG-CQT & 0.7255 & 0.8159 & 0.5576 & 0.6432 \\
		\midrule
		VGG-CQT-LE & 0.7404 & 0.8199 & 0.5859 & 0.6735 \\
		\midrule
		VGG-MFCC & 0.6451 & 0.7649 & 0.4571 & 0.5673 \\
		\midrule
		VGG-MFCC-LE & 0.6461 & 0.7685 & 0.4771 & 0.5938 \\
		\midrule
		ResNet-STFT & 0.7374 & 0.817 & 0.5248 & 0.6123 \\
		\midrule
		ResNet-STFT-LE & 0.745 & 0.8176 & 0.5456 & 0.6525 \\
		\midrule
		ResNet-CQT & 0.7221 & 0.8039 & 0.588 & \textbf{0.684} \\
		\midrule
		ResNet-CQT-LE & 0.7365 & 0.8257 & \textbf{0.5921} & 0.6796 \\
		\midrule
		ResNet-MFCC & 0.6553 & 0.7766 & 0.476 & 0.584 \\
		\midrule
		ResNet-MFCC-LE & 0.6675 & 0.7994 & 0.4866 & 0.6142 \\
		\bottomrule
	\end{tabular}%
	\label{tab:2}%
\end{table}%

\begin{table}[!htbp]
	\centering
	\caption{Classification performances on the DCASE2017 datasets us-ing multi-spectrogram fusion.}
	\begin{tabular}{cp{5em}cccc}
		\toprule
		\textbf{Dataset} & \textbf{Aggregated Features} & \textbf{VGG} & \textbf{VGG-LE} & \textbf{ResNet} & \textbf{ResNet-LE} \\
		\midrule
		Devel-Set & STFT+CQT & 0.8756 & \textbf{0.8865} & 0.8773 & 0.8852 \\
		\midrule
		Devel-Set & STFT+MFCC & 0.85  & 0.8592 & 0.8568 & 0.8747 \\
		\midrule
		Devel-Set & MFCC+CQT & 0.8364 & 0.8478 & 0.8508 & 0.8545 \\
		\midrule
		Eval-Set & STFT+CQT & 0.6914 & 0.7302 & 0.7654 & \textbf{0.7778} \\
		\midrule
		Eval-Set & STFT+MFCC & 0.6778 & 0.6988 & 0.7234 & 0.7321 \\
		\midrule
		Eval-Set & MFCC+CQT & 0.6222 & 0.6728 & 0.6907 & 0.6957 \\
		\bottomrule
	\end{tabular}%
	\label{tab:3}%
\end{table}%

\subsection{Evaluations on LITIS Rouen dataset}
We also evaluate our methods on the LITIS Rouen dataset. When using single spectrogram, the best sample-level accuracy and segment-level accuracy are 0.8901 and 0.9614 respectively, both achieved by ResNet-CQT-LE, as shown in Table~\ref{tab:4}. Table~\ref{tab:5} demonstrates the results of the multi-spectrogram fusions. With the aggregated features of STFT+CQT, the accuracies on the four models (VGG, VGG-LE, ResNet, ResNet-LE) are all above 0.97, outperforming the ones of all other aggregated features. The highest one obtained by the aggregated features of STFT+CQT is 0.9744, which is an improvement of 0.013 over the best accuracy using single spectrogram (i.e., 0.9614). It has again validated the effectiveness of the multi-spectrogram fusion. Moreover, our best accuracy has been very close to the current state-of-the-art accuracy (0.978) achieved in~\cite{Phan2017Audio}. It should be mentioned that no all fusions can improve the performances here. For example, the accuracies with the fusions of STFT+MFCC and MFCC+CQT are both less than 0.9614. These declines in accuracies have illustrated the importance of aggregated feature selection. For the comparison of basic models, there is no prevailing model for all aggregated features. In detail, for the aggregated features of STFT+CQT, the best basic model is ResNet, while the VGG-like models are better for the other two aggregated features.\\

\begin{table}[htbp]
	\centering
	\caption{Classification performances on the LITIS Rouen dataset using different single spectrogram.}
	\begin{tabular}{ccc}
		\toprule
		\textbf{Model} & \textbf{Sample-Level} & \textbf{Voting}\\
		\midrule
		VGG-STFT & 0.8504 & 0.9451 \\
		\midrule
		VGG-STFT-LE & 0.8566 & 0.9482 \\
		\midrule
		VGG-CQT & 0.8735 & 0.9564 \\
		\midrule
		VGG-CQT-LE & 0.8748 & 0.956 \\
		\midrule
		VGG-MFCC & 0.7114 & 0.9077 \\
		\midrule
		VGG-MFCC-LE & 0.7145 & 0.9117 \\
		\midrule
		ResNet-STFT & 0.8671 & 0.9458 \\
		\midrule
		ResNet-STFT-LE & 0.8687 & 0.947 \\
		\midrule
		ResNet-CQT & 0.8895 & 0.9588 \\
		\midrule
		ResNet-CQT-LE & \textbf{0.8901} & \textbf{0.9614} \\
		\midrule
		ResNet-MFCC & 0.7201 & 0.9051 \\
		\midrule
		ResNet-MFCC-LE & 0.7113 & 0.8997 \\
		\bottomrule
	\end{tabular}%
	\label{tab:4}%
\end{table}%

\begin{table}[htbp]
	\centering
	\caption{Classification performances on the LITIS Rouen dataset using multi-spectrogram fusion.}
	\begin{tabular}{ccccc}
		\toprule
		\textbf{Aggregated Features} & \textbf{VGG} & \textbf{VGG-LE} & \textbf{ResNet} & \textbf{ResNet-LE} \\
		\midrule
		STFT + CQT & 0.9703 & 0.9707 & \textbf{0.9744} & 0.9738 \\
		\midrule
		STFT + MFCC & 0.9502 & 0.9486 & 0.943 & 0.9476 \\
		\midrule
		MFCC+CQT & 0.9591 & 0.9555 & 0.9589 & 0.9586 \\
		\bottomrule
	\end{tabular}%
	\label{tab:5}%
\end{table}%

\begin{table}[htbp]
	\centering
	\caption{Accuracy comparison of different methods.}
	\begin{tabular}{p{6.5em}p{5em}p{5em}p{5em}}
		\toprule
		\textbf{Method} & \textbf{DCASE2017 Development Set} & \textbf{DCASE2017 Evaluation Set} & \textbf{LITIS Rouen Dataset} \\
		\midrule
		RNN-Fusion~\cite{Phan2017Improved}  & - & - & 0.978 \\
		\midrule
		CNN-Fusion6~\cite{Phan2017Improved}  & - & - & 0.966 \\
		\midrule
		Scene+Speech-LTE~\cite{Phan2016Label}  & - & - & 0.964 \\
		\midrule
		FisherHOG+ ProbSVM~\cite{Ye2015Acoustic}  & - & - & 0.96 \\
		\midrule
		CNN-LSTM-GAN~\cite{mun2017generative}  & 0.871 & 0.833\footnote{http://www.cs.tut.fi/sgn/arg/dcase2017/challenge/task-acoustic-scene-classification-results} & - \\
		\midrule
		Background Subtraction~\cite{Han2017Convolutional}  & 0.919 & $ 0.804^3 $ & - \\
		\midrule
		Mixup Multi-Channel~\cite{Xu2018Mixup}  & 0.872 & 0.767 & - \\
		\midrule
		Multi-temporal Resolution~\cite{Zhu2018Environmental}  & - & 0.747 & - \\
		\midrule
		LEW score fusion~\cite{hyderbuet}  & 0.887 & 0.741 & - \\
		\midrule
		Scalogram-CNN-GRNN~\cite{Ren2018Deep}  & 0.844 & 0.64  & - \\
		\midrule
		\textbf{Our method} & 0.8865 & 0.7778 & 0.9744 \\
		\bottomrule
	\end{tabular}%
	\label{tab:6}%
\end{table}%

\indent As shown in the Table~\ref{tab:5}, label expansion does not always work in the LITIS Rouen dataset. First, even it helps, the improvement gained is slight. Second, the accuracy would slightly decrease in some cases when it is applied. For example, with the aggregated features of STFT+CQT, the accuracy achieved by ResNet-LE is 0.9738 versus 0.9744 by ResNet. However, label expansion is found really work in DCASE2017 dataset. We suppose that the difference lies in the distributing characteristics of the datasets. When the clustered subsets in the dataset are separable, label expansion will play an effective role. On the contrary, it will fail when the boundaries of the subsets are ambiguous. To further testify its usefulness, we conduct experiments on another environmental sound dataset of ESC-50~\cite{Piczak2015ESC}. Using the aggregated features of STFT+CQT, the accuracy achieved by VGG is 0.7145 versus 0.7285 by VGG-LE. An improvement of 0.014 is gained by label expansion, which has proven its effectiveness.

\section{CONCLUSION}
\label{sec:4}
In this letter, we have proposed a CNN-based multi-spectrogram fusion framework for acoustic scene classification. The fusion of STFT and CQT spectrograms achieves the best performance for all datasets and CNN models in our experiments. Motivated by the inter-class similarities in the datasets, a label expansion method is proposed where super-class labels are constructed. The super-class labels are utilized to transform the CNN model into multitask learning paradigm. Moreover, the relations between the super-classes and the original scene classes are modeled as regularization constrains in the loss function. Label expansion takes effect on some datasets, depending on the distributing characteristics of the datasets. Generally, the proposed methods have achieved promising results on both the DCASE2017 and the LITIS Rouen datasets.

\ifCLASSOPTIONcaptionsoff
  \newpage
\fi

\newpage
\newpage


\begin{thebibliography}{10}
	\providecommand{\url}[1]{#1}
	\csname url@samestyle\endcsname
	\providecommand{\newblock}{\relax}
	\providecommand{\bibinfo}[2]{#2}
	\providecommand{\BIBentrySTDinterwordspacing}{\spaceskip=0pt\relax}
	\providecommand{\BIBentryALTinterwordstretchfactor}{4}
	\providecommand{\BIBentryALTinterwordspacing}{\spaceskip=\fontdimen2\font plus
		\BIBentryALTinterwordstretchfactor\fontdimen3\font minus
		\fontdimen4\font\relax}
	\providecommand{\BIBforeignlanguage}[2]{{%
			\expandafter\ifx\csname l@#1\endcsname\relax
			\typeout{** WARNING: IEEEtran.bst: No hyphenation pattern has been}%
			\typeout{** loaded for the language `#1'. Using the pattern for}%
			\typeout{** the default language instead.}%
			\else
			\language=\csname l@#1\endcsname
			\fi
			#2}}
	\providecommand{\BIBdecl}{\relax}
	\BIBdecl
	
	\bibitem{Barchiesi2015Acoustic}
	D.~Barchiesi, D.~Giannoulis, S.~Dan, and M.~D. Plumbley, ``Acoustic scene
	classification: Classifying environments from the sounds they produce,''
	\emph{IEEE Signal Processing Magazine}, vol.~32, no.~3, pp. 16--34, 2015.
	
	\bibitem{Yu2017Recent}
	D.~Yu and J.~Li, ``Recent progresses in deep learning based acoustic models,''
	\emph{IEEE/CAA Journal of Automatica Sinica}, vol.~4, no.~3, pp. 396--409,
	2017.
	
	\bibitem{Eghbal2016CP}
	H.~Eghbal-Zadeh, B.~Lehner, M.~Dorfer, and G.~Widmer, ``Cp-jku submissions for
	dcase-2016: a hybrid approach using binaural i-vectors and deep convolutional
	neural networks,'' 2016.
	
	\bibitem{Dehak2011Front}
	N.~Dehak, P.~J. Kenny, R.~Dehak, P.~Dumouchel, and P.~Ouellet, ``Front-end
	factor analysis for speaker verification,'' \emph{IEEE Transactions on Audio
		Speech \& Language Processing}, vol.~19, no.~4, pp. 788--798, 2011.
	
	\bibitem{Bae2016Acoustic}
	S.~H. Bae, I.~Choi, and N.~S. Kim, ``Acoustic scene classification using
	parallel combination of {LSTM} and {CNN},'' 2016.
	
	\bibitem{Graves1997Long}
	A.~Graves, ``Long short-term memory,'' \emph{Neural Computation}, vol.~9,
	no.~8, pp. 1735--1780, 1997.
	
	\bibitem{mun2017generative}
	S.~Mun, S.~Park, D.~K. Han, and H.~Ko, ``Generative adversarial network based
	acoustic scene training set augmentation and selection using svm
	hyper-plane,'' \emph{Proc. DCASE}, pp. 93--97, 2017.
	
	\bibitem{Goodfellow2014Generative}
	I.~J. Goodfellow, J.~Pouget-Abadie, M.~Mirza, B.~Xu, D.~Warde-Farley, S.~Ozair,
	A.~Courville, and Y.~Bengio, ``Generative adversarial networks,''
	\emph{Advances in Neural Information Processing Systems}, vol.~3, pp.
	2672--2680, 2014.
	
	\bibitem{Ren2018Deep}
	Z.~Ren, K.~Qian, Z.~Zhang, V.~Pandit, A.~Baird, and B.~Schuller, ``Deep
	scalogram representations for acoustic scene classification,'' \emph{IEEE/CAA
		Journal of Automatica Sinica}, vol.~5, no.~3, 2018.
	
	\bibitem{Chung2014Empirical}
	J.~Chung, C.~Gulcehre, K.~H. Cho, and Y.~Bengio, ``Empirical evaluation of
	gated recurrent neural networks on sequence modeling,'' \emph{Eprint Arxiv},
	2014.
	
	\bibitem{Zhang2016Embedding}
	X.~Zhang, F.~Zhou, Y.~Lin, and S.~Zhang, ``Embedding label structures for
	fine-grained feature representation,'' in \emph{IEEE Conference on Computer
		Vision and Pattern Recognition}, 2016, pp. 1114--1123.
	
	\bibitem{Han2017Breast}
	Z.~Han, B.~Wei, Y.~Zheng, Y.~Yin, K.~Li, and S.~Li, ``Breast cancer
	multi-classification from histopathological images with structured deep
	learning model,'' \emph{Scientific Reports}, vol.~7, no.~1, p. 4172, 2017.
	
	\bibitem{Nawab1988Short}
	S.~H. Nawab, ``Short-time fourier transform,'' in \emph{Advanced Topics in
		Signal Processing}, 1988, pp. 289--337.
	
	\bibitem{Brown1992An}
	J.~C. Brown and M.~S. Puckette, ``An efficient algorithm for the calculation of
	a constant q transform,'' \emph{Journal of the Acoustical Society of
		America}, vol.~92, no.~5, p. 2698, 1992.
	
	\bibitem{Logan2000Mel}
	B.~Logan, ``Mel frequency cepstral coefficients for music modeling,''
	\emph{Proc of Ismir}, 2000.
	
	\bibitem{He2015Deep}
	K.~He, X.~Zhang, S.~Ren, and J.~Sun, ``Deep residual learning for image
	recognition,'' pp. 770--778, 2015.
	
	\bibitem{Szegedy2014Going}
	C.~Szegedy, W.~Liu, Y.~Jia, P.~Sermanet, S.~Reed, D.~Anguelov, D.~Erhan,
	V.~Vanhoucke, and A.~Rabinovich, ``Going deeper with convolutions,'' pp.
	1--9, 2014.
	
	\bibitem{Hershey2017CNN}
	S.~Hershey, S.~Chaudhuri, D.~P.~W. Ellis, J.~F. Gemmeke, A.~Jansen, R.~C.
	Moore, M.~Plakal, D.~Platt, R.~A. Saurous, and B.~Seybold, ``Cnn
	architectures for large-scale audio classification,'' in \emph{IEEE
		International Conference on Acoustics, Speech and Signal Processing}, 2017,
	pp. 131--135.
	
	\bibitem{Sejdi2009Time}
	E.~Sejdić, I.~Djurović, and J.~Jiang, \emph{Time--frequency feature
		representation using energy concentration: An overview of recent
		advances}.\hskip 1em plus 0.5em minus 0.4em\relax Academic Press, Inc., 2009.
	
	\bibitem{Boashash2015Time}
	B.~Boashash, N.~A. Khan, and T.~Ben-Jabeur, \emph{Time-frequency features for
		pattern recognition using high-resolution TFDs}.\hskip 1em plus 0.5em minus
	0.4em\relax Academic Press, Inc., 2015.
	
	\bibitem{Battaglino2016Acoustic}
	D.~Battaglino, ``Acoustic scene classification using convolutional neural
	networks,'' 2016.
	
	\bibitem{Lidy2016CQT}
	T.~Lidy and A.~Schindler, ``Cqt-based convolutional neural networks for audio
	scene classification,'' in \emph{Detection and Classification of Acoustic
		Scenes and Events 2016 Workshop}, 2016.
	
	\bibitem{weiping2017acoustic}
	Z.~Weiping, Y.~Jiantao, X.~Xiaotao, L.~Xiangtao, and P.~Shaohu, ``Acoustic
	scene classification using deep convolutional neural network and multiple
	spectrograms fusion,'' in \emph{Detection and Classification of Acoustic
		Scenes and Events 2017 Workshop (DCASE2017)}, 2017.
	
	\bibitem{Ng2001On}
	A.~Y. Ng, M.~I. Jordan, and Y.~Weiss, ``On spectral clustering: analysis and an
	algorithm,'' \emph{Proc Nips}, vol.~14, pp. 849--856, 2001.
	
	\bibitem{Xie2015Hyper}
	S.~Xie, T.~Yang, X.~Wang, and Y.~Lin, ``Hyper-class augmented and regularized
	deep learning for fine-grained image classification,'' in \emph{IEEE
		Conference on Computer Vision and Pattern Recognition}, 2015, pp. 2645--2654.
	
	\bibitem{Gopal2013Recursive}
	S.~Gopal and Y.~Yang, ``Recursive regularization for large-scale classification
	with hierarchical and graphical dependencies,'' in \emph{ACM SIGKDD
		International Conference on Knowledge Discovery and Data Mining}, 2013, pp.
	257--265.
	
	\bibitem{Rakotomamonjy2017Histogram}
	A.~Rakotomamonjy and G.~Gasso, ``Histogram of gradients of time–frequency
	representations for audio scene classification,'' \emph{IEEE/ACM Transactions
		on Audio Speech \& Language Processing}, vol.~23, no.~1, pp. 142--153, 2017.
	
	\bibitem{Abadi2015TensorFlow}
	M.~Abadi, A.~Agarwal, P.~Barham, E.~Brevdo, Z.~Chen, C.~Citro, G.~S. Corrado,
	A.~Davis, J.~Dean, and M.~Devin, ``Tensorflow: Large-scale machine learning
	on heterogeneous distributed systems,'' 2015.
	
	\bibitem{Kingma2014Adam}
	D.~Kingma and J.~Ba, ``Adam: A method for stochastic optimization,''
	\emph{Computer Science}, 2014.
	
	\bibitem{Phan2017Audio}
	H.~Phan, P.~Koch, F.~Katzberg, M.~Maass, R.~Mazur, and A.~Mertins, ``Audio
	scene classification with deep recurrent neural networks,'' 2017.
	
	\bibitem{Piczak2015ESC}
	K.~J. Piczak, ``Esc: Dataset for environmental sound classification,'' in
	\emph{ACM International Conference on Multimedia}, 2015, pp. 1015--1018.
	
	\bibitem{Phan2017Improved}
	H.~Phan, L.~Hertel, M.~Maass, P.~Koch, R.~Mazur, and A.~Mertins, ``Improved
	audio scene classification based on label-tree embeddings and convolutional
	neural networks,'' \emph{IEEE/ACM Transactions on Audio Speech \& Language
		Processing}, vol.~25, no.~6, pp. 1278--1290, 2017.
	
	\bibitem{Phan2016Label}
	H.~Phan, L.~Hertel, M.~Maass, P.~Koch, and A.~Mertins, ``Label tree embeddings
	for acoustic scene classification,'' in \emph{ACM on Multimedia Conference},
	2016, pp. 486--490.
	
	\bibitem{Ye2015Acoustic}
	J.~Ye, T.~Kobayashi, M.~Murakawa, and T.~Higuchi, ``Acoustic scene
	classification based on sound textures and events,'' in \emph{ACM
		International Conference on Multimedia}, 2015, pp. 1291--1294.
	
	\bibitem{Han2017Convolutional}
	Y.~Han, J.~Park, and K.~Lee, ``Convolutional neural networks with binaural
	represntations and background subtraction for acoustic scene
	classification,'' in \emph{Detection and Classification of Acoustic Scenes
		and Events}, 2017.
	
	\bibitem{Xu2018Mixup}
	K.~Xu, D.~Feng, H.~Mi, B.~Zhu, D.~Wang, L.~Zhang, H.~Cai, and S.~Liu,
	``Mixup-based acoustic scene classification using multi-channel convolutional
	neural network,'' 2018.
	
	\bibitem{Zhu2018Environmental}
	B.~Zhu, K.~Xu, D.~Wang, L.~Zhang, B.~Li, and Y.~Peng, ``Environmental sound
	classification based on multi-temporal resolution convolutional neural
	network combining with multi-level features,'' 2018.
	
	\bibitem{hyderbuet}
	R.~Hyder, S.~Ghaffarzadegan, Z.~Feng, and T.~Hasan, ``Buet bosch consortium
	(b2c) acoustic scene classification systems for dcase 2017 challenge.''
	
\end{thebibliography}
\end{document}